\documentclass[runningheads]{llncs}
\usepackage[T1]{fontenc}
\usepackage{graphicx}
\usepackage{booktabs}
\usepackage{amsmath}
\usepackage{amsfonts}
\usepackage{multirow}
\usepackage{makecell}
\usepackage{hyperref}
\usepackage{xcolor}

\newcommand{\new}[1]{#1}
\newcommand\blfootnote[1]{\begingroup\renewcommand\thefootnote{}\footnotetext{#1}\endgroup}
\begin{document}
\title{Controllable Generation of Diverse Dermatological Imagery for Fair and Efficient Malignancy Classification}
\titlerunning{cgDDI}
\author{Héctor Carrión\inst{1} \and Narges Norouzi\inst{2}}
\authorrunning{Carrión et al.}
  \institute{University of California, Santa Cruz, USA \and
University of California, Berkeley, USA \\
\email{hcarrion@ucsc.edu, norouzi@berkeley.edu}}

\maketitle
\blfootnote{Author's accepted manuscript. Accepted at MICCAI 2026 (Springer LNCS). The final authenticated version will be linked here upon publication.}
\begin{abstract}
\setlength{\emergencystretch}{3em}%
Accurate dermatological diagnosis naturally necessitates equitable performance across diverse populations, yet a systematic lack of expertly annotated images, especially for underrepresented skin tones and rare diseases, impedes progress toward measurably fair methods. We introduce \textbf{cgDDI} (\textbf{C}ontrollable \textbf{G}eneration of \textbf{D}iverse \textbf{D}ermatological \textbf{I}magery), a hybrid framework that (1)~synthesizes realistic healthy skin samples without disturbing other input properties, (2)~maps single-sample rare lesions onto novel skin-tones and locations non-parametrically, and (3)~allows for efficient parametric generation with as few as 10~training samples. The framework supports both human and automated segmentation masking, enabling scalability to datasets without pre-made lesion masks. We grow a 656-image dataset by more than $400\times$ and validate across two datasets: biopsy-confirmed Diverse Dermatology Images (DDI) and expert-verified Fitzpatrick17k (F17k). On the DDI benchmark, we achieve malignancy classification accuracy of $86.4\%$ under syn\-thet\-ic-only training and $90.9\%$ state-of-the-art performance with real data fine-tuning, alongside leading fairness metrics. Cross-dataset experiments show $+13.9\%$ accuracy improvements on unseen F17k data despite minimal disease overlap. We openly release 266k+ synthetic images, code, and generative models to further support fairness research at \url{https://github.com/hectorcarrion/ControllableGenDDI}.

\end{abstract}

\section{Introduction}
\label{sec:intro}


Skin diseases affect millions globally, with expert diagnosis accuracy being measurably lower for darker-skinned populations, even if the physicians originate from diverse backgrounds~\cite{groh-2024}. Furthermore, over 3~billion people lack access to adequate dermatological care, especially those in impoverished communities~\cite{coustasse_2019_use}. Early detection of skin cancers significantly increases survival rates~\cite{balch-2009}, yet Artificial Intelligence (AI) systems trained on biased data sources risk exacerbating disparities. A survey of 70 dermatological AI studies found fewer than 25\% included ethnicity and only 10\% included skin-tone descriptors~\cite{daneshjou_2021_lack}. Recent generative approaches addressing this issue require large or private training sets~\cite{internal_1,ktena-2024}, do not cover the full skin-tone spectrum~\cite{wang-2024}, or ignore extremely rare diseases~\cite{internal_2}. We present cgDDI, a novel hybrid generation framework (Fig.~\ref{fig:framework}) addressing these limitations through complementary parametric and non-parametric approaches. Our contributions are:


\begin{figure}[t]
  \centering
  \includegraphics[width=\textwidth]{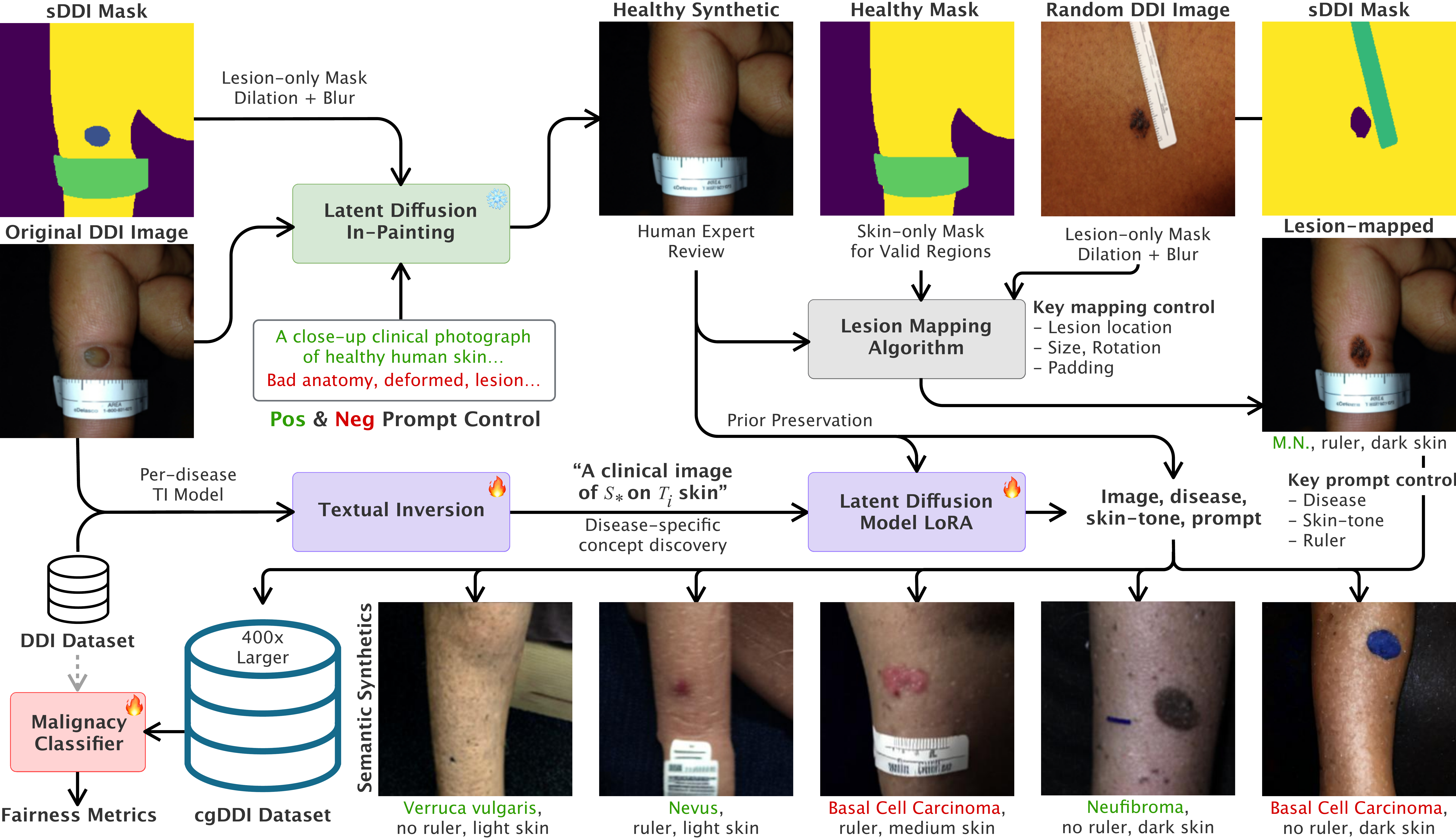}
   \caption{\textbf{cgDDI framework.} Original images, masks, and prompts produce \textbf{healthy synthetics}. These serve as targets for \textbf{lesion-mapped synthetics}, as prior-preservation anchors, and as semantic prompts. Disease-specific concepts are learned via textual inversion and used to fine-tune latent diffusion models from which \textbf{semantic synthetics} are sampled. The aggregated data trains \textbf{fair classifiers}.}
   \label{fig:framework}
\end{figure}

\begin{enumerate}
    \item \textbf{cgDDI Framework:} A controllable method combining (a)~latent diffusion inpainting for pixel-perfect healthy skin synthesis, (b)~non-parametric lesion mapping enabling \new{\emph{single-sample} disease augmentation where parametric methods struggle}, and (c)~efficient parametric generation via textual inversion and Low Rank Adaptation (LoRA) regularized \new{by, to our knowledge, the first use of prior-preservation loss (PPL) anchored on healthy images}.
    \item \textbf{cgDDI Dataset:} 266,136 skin-tone-balanced synthetic images (healthy, lesion-mapped, semantic) with fairness labels, openly released, \new{including 309 pixel-perfect in-clinical-distribution healthy synthetics, a resource not present in prior work} (Table~\ref{tab:datasets}).
    \item \textbf{Classification and Fairness:} Malignancy classification on the DDI benchmark reaches SOTA accuracy ($90.9\%$, up $3.5\%$) \new{under full Fitzpatrick I--VI coverage} including rich fairness metrics for both discriminative and generative results. Cross-dataset validation on F17k demonstrates generalizability with $+4.7\%$ intra- and $+13.9\%$ cross-dataset accuracy improvements.
\end{enumerate}

\begin{table}[t]
  \centering
  \caption{Comparison of synthetic dermatological datasets across training data, fairness indicators, skin-tone coverage, number of learned diseases, \new{single-view augmentation support and whether the resulting datasets are published openly.}}\label{tab:datasets}
  \setlength\tabcolsep{3pt}
  \fontsize{8}{9.5}\selectfont
  \begin{tabular}{lccccccc}
    \toprule
    Method
      & \makecell{Training\\Data}
      & \makecell{Fairness\\Metrics}
      & \makecell{FST\\Cov.}
      & \makecell{Ctrl.\\Diseases}
      & \makecell{Total\\Size}
      & \new{\makecell{Single-\\view}}
      & \makecell{Open\\Data} \\
    \midrule
    Sagers et al. (2022) \cite{internal_2}   & F17k    & FST Acc. & I--VI  & 3  & 192  & \new{--} & -- \\
    Sagers et al. (2023) \cite{sagers-2023}  & F17k, DDI & FST Acc. & I--VI & 9 & 459k & \new{--} & \checkmark \\
    Akrout et al. (2024) \cite{internal_1}   & Private & None     & None   & 6  & 180k & \new{--} & -- \\
    Ktena et al. (2024) \cite{ktena-2024}   & Private & FST Gap  & I--VI  & 27 & 50k  & \new{--} & -- \\
    Wang et al. (2024) \cite{wang-2024}    & F17k    & FST Acc. & I--II, V--VI & 7 & 7.6k & \new{--} & -- \\
    \midrule
    \textbf{cgDDI (ours)} & DDI   & Multiple & I--VI  & 13 & 266k & \new{\checkmark} & \checkmark \\
    \bottomrule
  \end{tabular}
\end{table}

\section{Related Work}
\label{sec:related}

The F17k~\cite{groh_2021_evaluating} dataset is widely used toward fairness research thanks to its large 17,000 sample size, however, it suffers from $3.6\times$ skin-tone imbalance and $>30\%$ label noise from non-expert annotations~\cite{daneshjou_2022_checklist}. Recently an expert-verified F17k subset~\cite{groh-2024} has been released but it is much smaller in scope (364~samples). The DDI dataset~\cite{daneshjou_2022_disparities} is relatively larger (656~samples) fully biopsy-confirmed with dermatologist-verified Fitzpatrick-scale labels, and mostly balanced across skin tones, with sDDI~\cite{carrion-2023} providing segmentation masks. For these reasons, we train our main models on DDI but evaluate cross-dataset with expert-verified F17k. Recent malignancy classification fairness work has advanced through contrastive disentanglement~\cite{du_2023_fairdisco} and patch alignment~\cite{aayushman-2024}, establishing the DDI benchmark for both accuracy and fairness metrics which we compare against.

Diffusion models (DMs)~\cite{dhariwal_2021_diffusion} pre-trained for text-conditioned generation have shown controllability in dermatology~\cite{sagers-2023}, with textual inversion~\cite{gal2023an} improving control~\cite{wang-2024}. However, existing frameworks do not cover rare disease conditions with low frequencies, may rely on noisy or private training data, may not release generated outputs, several have incomplete skin-tone coverage and none report rich fairness metrics. We survey recent synthetic datasets in Table~\ref{tab:datasets}. Our work addresses each of these gaps.

\section{cgDDI Framework}
\label{sec:method}

cgDDI generates three complementary types of synthetic dermatological imagery through a sequential pipeline. We first consolidate DDI (Sec.~\ref{sec:basedata}), develop a latent diffusion inpainting algorithm to create healthy samples (Sec.~\ref{sec:inpainting}), which then serve as canvas for non-parametric lesion mapping (Sec.~\ref{sec:lesion_mapping}) and as prior-preservation anchors for parametric semantic generation (Sec.~\ref{sec:parametric}). We generate 309 healthy, 80,427 lesion-mapped, and 185,400 semantic synthetic images.


\subsection{Data Pre-processing}
\label{sec:basedata}

DDI contains 656~samples (171 malignant, 485 benign). We consolidate 78 original disease labels into 65 categories based on histopathological similarity as in previous work \cite{tschandl_2018_the}, yielding 25 single-observation diseases, 27 diseases between 2 and 10~observations, and 13 diseases with $>$10~observations.

\subsection{Healthy Synthesis via Latent Diffusion Inpainting}
\label{sec:inpainting}

\begin{figure}[t]
  \centering
  \includegraphics[width=\textwidth]{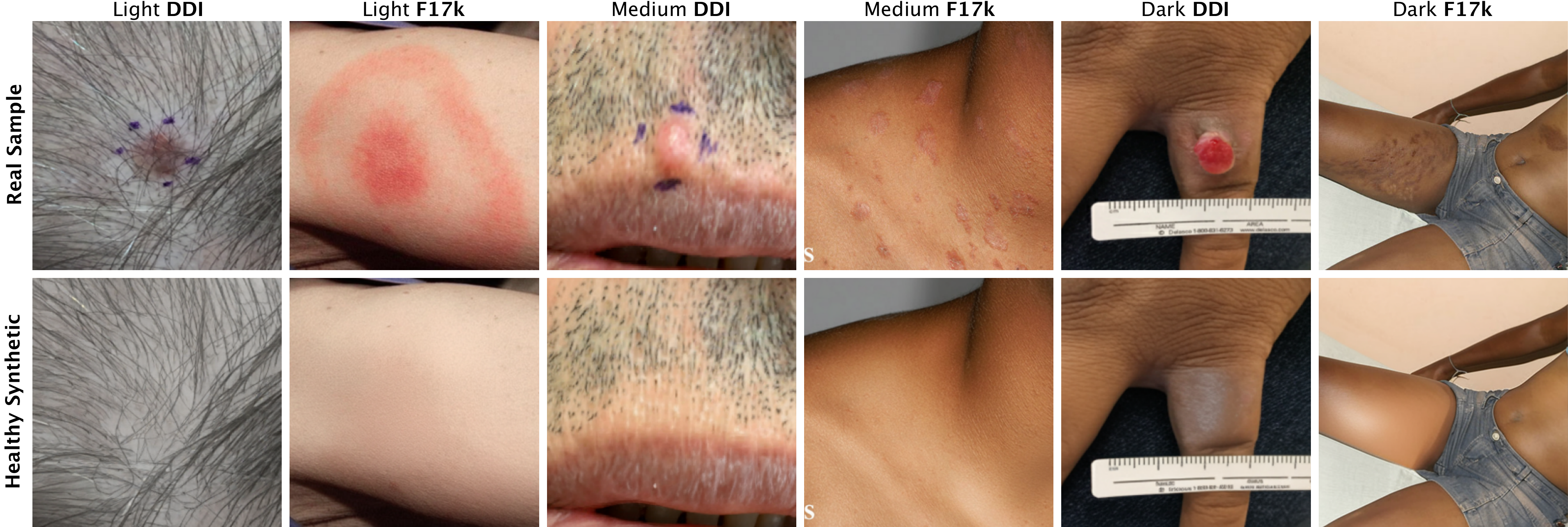}
   \caption{\textbf{Healthy synthetic imagery.} Our inpainting removes target lesions (and markers), producing lesion-less reconstruction robust to hair, morphology, and body location. These samples are later used for lesion mapping and in-distribution prior preservation. Results shown for DDI (human masking) and F17k (algorithmic masking).}
   \label{fig:healthy}
\end{figure}

To our knowledge, a dataset providing dermatologist-verified healthy skin imagery collected analogously to diseased samples does not exist. We create it by inpainting lesion regions using a UNet denoiser (1.22B parameters) and MoVQGAN decoder~\cite{razzhigaev-2023}, guided by positive (\textit{``healthy, smooth, normal human skin''}) and negative (\textit{``lesion, hole, transparent, eye''}) prompts. Segmentation masks delineate the inpainting regions; we apply dilation and Gaussian blur to smooth mask boundaries. Given 334 masked sDDI~\cite{carrion-2023} inputs, we retain 309 healthy synthetics after human review (7\% discard rate for generative artifacts). Results are shown in Fig.~\ref{fig:healthy}.

While we leverage human-made masks for DDI, our framework is compatible with algorithmic masking. We demonstrate this by masking F17k~\cite{groh-2024} via SAMv3~\cite{carion2025sam3segmentconcepts} for fully automated mask generation. SAMv3 successfully segments lesions across skin tones without dataset-specific training which our framework inputs directly, confirming generalizability and scaling to datasets without pre-made annotations. This is shown on Fig.~\ref{fig:healthy}. \new{We note that SAMv3 masks can be noisy for medical images, and thus meet our discard criteria more often than human-made masks. Failure-mode examples are discussed on our repository.}

\subsection{Non-Parametric Lesion Mapping}
\label{sec:lesion_mapping}

For extremely rare diseases where parametric learning is infeasible, we transplant real lesions onto healthy canvases. Our algorithm leverages segmentation masks to identify valid skin regions, and follows padding constraints (e.g. minimum 10~pixels from skin edges) to avoid placing lesions on unrealistic positions. Given 309 healthy images and 334 donor masks, we generate 80,427 lesion-mapped samples (22\% discarded by padding constraints). This non-parametric approach enables augmentation from single-sample diseases zero-shot.

\subsection{Parametric Semantic Generation}
\label{sec:parametric}

For diseases with $\geq$10 samples, we learn disease-specific tokens through textual inversion~\cite{gal2023an}, then fine-tune the latent DM backbone~\cite{rombach-2022} via LoRA~\cite{hu2022lora}. Critically, we are the first in dermatological generation to employ Prior Preservation Loss (PPL)~\cite{ruiz-2023}, using our healthy synthetics as an in-distribution regularization set. PPL addresses two issues in DM fine-tuning: semantic drift (forgetting class-level knowledge) and reduced output diversity, enabling more faithful generation than textual inversion alone~\cite{zeng-2024}. Our healthy synthetics are uniquely suited for this role as they share the clinical imaging conditions and verified skin-tone labels of the training data.

At generation time, for healthy images $H{=}\{h_i\}_{i=1}^{309}$, diseases $D{=}\{d_j\}_{j=1}^{40}$, and skin tones $S{=}\{s_k\}_{k=1}^{3}$, each triple $(h_i, d_j, s_k)$ produces $R{=}5$ samples:
\begin{equation}
    x_{i,j,k}^{(r)}
    = f_{\theta,j}\bigl(
         h_i,\;
         \textit{``An image of $S_{*,j}$ on a $s_k$-toned individual''}
         \;;\;\alpha,\,\beta,\,t
      \bigr)
\end{equation}
where $f_{\theta,j}$ is the disease-specific DM, $S_{*,j}$ is the learned token, $\alpha$ is the conditioning strength, $\beta$ the guidance scale, and $t$ the inference steps. This yields $|H|{\times}|D|{\times}|S|{\times}R = 185{,}400$ semantic synthetics, balanced at 1,545 per skin tone per disease. We find $\sim$10 samples sufficient for viable generation quality, but release all synthesized images in our dataset repository for further study.

\begin{figure}[t]
  \centering
  \includegraphics[width=\textwidth]{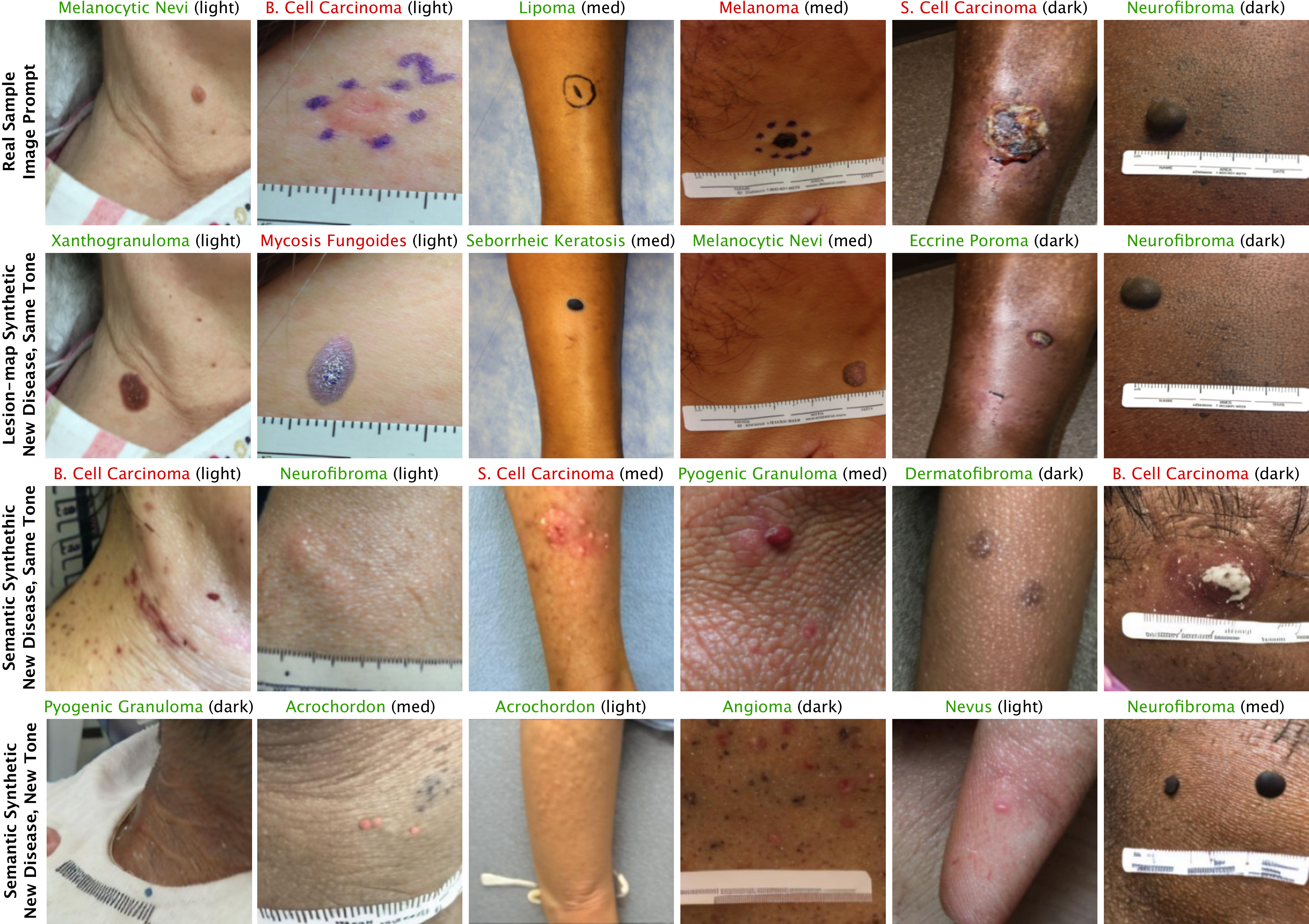}
   \caption{\textbf{cgDDI samples.} Row~1: real images used as prompts. Row~2: lesion from a donor transplanted onto the prompt image. Rows~3--4: semantic synthetics conditioned on the prompt, a target disease (\textcolor{red}{malignant}/\textcolor{green}{benign}), and target skin tone (light, medium, dark).}
   \label{fig:samples}
\end{figure}

\section{Experiments}
\label{sec:experiments}

\subsection{Setup and Metrics}
\label{sec:setup}

We adopt the PatchAlign~\cite{aayushman-2024} classifier (ViT-B/16) and evaluation protocol for direct comparison with prior state-of-the-art including their five-fold cross-validation with the same seeds. \new{The PatchAlign training pipeline applies standard augmentations (crops, rotations, color jitter and flips); cgDDI gains are reported \emph{on top of} this stack and operate orthogonally along attributes such as skin tone and lesion morphology.} We report three fairness metrics: \textbf{PQD}~(Predictive Quality Disparity), the ratio of worst-to-best skin-tone accuracy; \textbf{DPM}~(Demographic Parity), measuring positive-prediction-rate consistency; and \textbf{EOM}~(Equality of Opportunity), measuring true-positive-rate consistency and identified as the most important metric by~\cite{aayushman-2024}. Please see PatchAlign~\cite{aayushman-2024} for full formula definitions. We additionally evaluate generative quality via FID~\cite{heusel-2017}, KID~\cite{binkowski-2018}, and LPIPS~\cite{zhang2018perceptual} between cgDDI and held-out real images, stratified by skin tone. \new{Test-set details per dataset: DDI includes 131 test samples per fold (42 Light I--II, 48 Medium III--IV, 41 Dark V--VI); F17k 15\% hold-out includes 55 test-set samples (22 Light, 17 Medium, 16 Dark). We recognize that in ideal conditions these test sets should be larger, and encourage the community to collect more sets of biopsy confirmed observations alongside accurate skin-tone labels.}



\subsection{Malignancy Classification}
\label{sec:ddi_cls}

We run two experiments: \textit{Exp.~1} trains purely on cgDDI synthetics; \textit{Exp.~2} trains on synthetics then fine-tunes on real DDI data following~\cite{aayushman-2024}. Both are evaluated on held-out real DDI test sets with leakage prevention (i.e. excluding training synthetics conditioned on downstream test images).

\begin{table}[t]
  \caption{\textbf{DDI classification and fairness.} R.~= real, S.~= synthetic. Bold = best, underline = second best. Our method achieves SOTA accuracy and leading EOM.}\label{tab:ddi_results}
  \centering
  \fontsize{8}{9.5}\selectfont
  \setlength\tabcolsep{3pt}
  \begin{tabular}{lccccccc}
    \toprule
      & \multicolumn{4}{c}{\textbf{Accuracy (\%)} $\pm$ Std}
      & \multicolumn{3}{c}{\textbf{Fairness} $\pm$ Std} \\
    \cmidrule(lr){2-5} \cmidrule(lr){6-8}
    Method
      & Mean & Light & Med. & Dark
      & PQD  & DPM   & EOM \\
    \midrule
    Baseline (R.)
      & $82.4{\pm}1.5$ & $83.3{\pm}1.0$ & $74.6{\pm}5.7$ & $89.7{\pm}2.2$
      & $77.0{\pm}1.9$ & \underline{$75.2$}${\pm}13$ & $58.7{\pm}4.3$ \\
    FairDisCo~\cite{du_2023_fairdisco}
      & $83.8{\pm}0.4$ & $88.6{\pm}0.1$ & $71.7{\pm}2.2$ & $92.0{\pm}2.8$
      & $78.0{\pm}4.5$ & $72.8{\pm}12$ & $63.7{\pm}3.5$ \\
    PatchAlign~\cite{aayushman-2024}
      & \underline{$87.4$}${\pm}1.2$ & \underline{$89.6$}${\pm}2.6$
      & $80.3{\pm}5.7$ & \underline{$92.3$}${\pm}1.3$
      & $86.9{\pm}6.1$ & $74.9{\pm}12$ & $69.6{\pm}1.7$ \\
    \midrule
    Exp.~1 (S.)
      & $86.4{\pm}1.0$ & $88.9{\pm}1.5$ & \underline{$84.1$}${\pm}2.6$ & $86.0{\pm}1.8$
      & $\mathbf{94.6}{\pm}3.1$ & $\mathbf{82.0}{\pm}9.7$ & \underline{$81.9$}${\pm}2.8$ \\
    Exp.~2 (S.+R.)
      & $\mathbf{90.9}{\pm}1.3$ & $\mathbf{93.3}{\pm}2.2$ & $\mathbf{86.4}{\pm}4.1$ & $\mathbf{93.0}{\pm}1.0$
      & \underline{$92.5$}${\pm}2.5$ & $68.8{\pm}11$ & $\mathbf{86.6}{\pm}1.9$ \\
    \bottomrule
  \end{tabular}
\end{table}

Results are shown in Table~\ref{tab:ddi_results}. Exp.~1 achieves competitive accuracy ($86.4\%$) while substantially improving all fairness metrics over prior methods. Exp.~2 achieves SOTA accuracy ($90.9\%$) across all skin tones. EOM raises from $69.6$ to $86.6$, improving fairness significantly. Stratifying by disease rarity (Table~\ref{tab:ddi_rarity}), synthetic-only performance remains competitive, including under very rare conditions (1--2 samples: $83.3\%$), supporting lesion mapping for single-sample augmentation. Fine-tuning on real data primarily benefits common diseases ($>$10 samples), consistent with the real-life scarcity of rare samples.

\begin{table}[t]
  \centering
  \caption{\textbf{Disease rarity performance.} Test-set accuracy stratified by the number of original real data observations per disease.}\label{tab:ddi_rarity}
  \fontsize{8}{9.5}\selectfont
  \setlength\tabcolsep{4pt}
  \begin{tabular}{lccc}
    \toprule
    Disease Rarity & \makecell{Cases in\\test set} & \makecell{Exp.~1 (S.)\\Accuracy (\%)} & \makecell{Exp.~2 (S.+R.)\\Accuracy (\%)} \\
    \midrule
    Common ($>$10 samples) & 107 & 85.05 & 91.59 \\
    Rare (3--10 samples) & 19 & 94.74 & 89.47 \\
    Very rare (1--2 samples) & 6 & 83.33 & 83.33 \\
    \bottomrule
  \end{tabular}
\end{table}


\subsection{Cross-Dataset Validation}
\label{sec:cross}

To demonstrate generalizability beyond DDI, we process the expert-verified F17k subset~\cite{groh-2024} (364~samples) through our full pipeline using SAMv3 automated masking, producing 46 healthy, 1,124 lesion-mapped, and 5,520 semantic synthetics after discard criteria.

\paragraph{Intra-Dataset (F17k $\rightarrow$ F17k).} Table~\ref{tab:cross} (top) shows that training on F17k synthetics then fine-tuning on real data achieves $90.7\%$ accuracy ($+4.7\%$ over baseline) with the highest PQD, confirming cgDDI effectiveness on a second dataset.

\paragraph{Cross-Dataset Transfer.}
We synthesize inter-dataset data by mapping lesions and generating semantics across DDI and F17k, producing 23,216 lesion-mapped and 46,050 semantic cross-dataset synthetics. Table~\ref{tab:cross} (bottom) shows transfer results. Most notably, DDI-trained synthetics improve F17k accuracy by $+13.9\%$ (from $60.5\%$ to $74.4\%$) despite only one shared disease condition (cutaneous T-cell lymphoma). Aggregated mixed-dataset training achieves up to $93.0\%$ on F17k and $86.7\%$ on DDI, with synthetics consistently improving over baselines.

\begin{table}[t]
  \caption{\textbf{Cross-dataset classification results.} Intra-dataset F17k validation (top), cross-dataset transfer (middle), and aggregated mixed-dataset training (bottom). Baselines are trained on real data only for the corresponding setting.}\label{tab:cross}
  \centering
  \fontsize{8}{9.5}\selectfont
  \setlength\tabcolsep{3pt}
  \begin{tabular}{llccccccc}
    \toprule
    Setting & Method & Acc & Light & Med. & Dark & PQD & DPM & EOM \\
    \midrule
    \multirow{3}{*}{\makecell[l]{F17k\\$\rightarrow$ F17k}}
      & Baseline     & 86.0 & 86.7 & 82.4 & \textbf{90.9} & 0.906 & \textbf{0.455} & 0.500 \\
      \cmidrule(lr){2-9}
      & Synth Only   & 88.4 & 86.7 & \textbf{94.1} & 81.8 & 0.869 & 0.441 & 0.500 \\
      & Synth + Real & \textbf{90.7} & 86.7 & \textbf{94.1} & \textbf{90.9} & \textbf{0.921} & 0.441 & 0.500 \\
    \midrule
    \multirow{3}{*}{\makecell[l]{F17k\\$\rightarrow$ DDI}}
      & Baseline     & \textbf{79.6} & \textbf{64.3} & \textbf{82.2} & \textbf{87.5} & 0.735 & 0.500 & 0.500 \\
      \cmidrule(lr){2-9}
      & Synth Only   & 74.3 & 57.1 & 77.8 & 82.5 & 0.693 & 0.604 & 0.440 \\
      & Synth + Real & 75.2 & \textbf{64.3} & 80.0 & 77.5 & \textbf{0.804} & \textbf{0.678} & \textbf{0.703} \\
    \midrule
    \multirow{3}{*}{\makecell[l]{DDI\\$\rightarrow$ F17k}}
      & Baseline     & 60.5 & 66.7 & 47.1 & 72.7 & 0.647 & 0.515 & 0.526 \\
      \cmidrule(lr){2-9}
      & Synth Only   & \textbf{74.4} & \textbf{80.0} & \textbf{70.6} & 72.7 & \textbf{0.882} & 0.556 & \textbf{0.613} \\
      & Synth + Real & 69.8 & 73.3 & 76.5 & 54.5 & 0.713 & \textbf{0.664} & 0.388 \\
    \midrule
    \multirow{3}{*}{\makecell[l]{Mix\\$\rightarrow$ F17k}}
      & Baseline     & 86.0 & 86.7 & 88.2 & 81.8 & \textbf{0.927} & \textbf{0.471} & 0.500 \\
      \cmidrule(lr){2-9}
      & Synth Only   & 86.1 & \textbf{93.3} & 88.2 & 72.7 & 0.779 & 0.378 & \textbf{0.750} \\
      & Synth + Real & \textbf{93.0} & \textbf{93.3} & 88.2 & \textbf{100.0} & 0.882 & 0.378 & 0.500 \\
    \midrule
    \multirow{3}{*}{\makecell[l]{Mix\\$\rightarrow$ DDI}}
      & Baseline     & 83.2 & 75.0 & 82.2 & 90.0 & 0.833 & \textbf{0.679} & 0.784 \\
      \cmidrule(lr){2-9}
      & Synth Only   & 79.7 & 71.4 & 84.4 & 80.0 & 0.846 & 0.436 & \textbf{0.833} \\
      & Synth + Real & \textbf{86.7} & \textbf{82.1} & \textbf{84.4} & \textbf{92.5} & \textbf{0.888} & 0.600 & 0.750 \\
    \bottomrule
  \end{tabular}
\end{table}

\subsection{Synthesis Ablation Study}
\label{sec:ablation}

We evaluate the individual contribution of each generation method by training classifiers on different subsets of cgDDI (Table~\ref{tab:ablation}). Training solely on healthy and lesion-mapped synthetics slightly reduces overall accuracy compared to real DDI ($-1.2\%$) but improves medium skin-tone performance, likely due to the limited morphological diversity in non-parametric outputs. Semantic synthetics alone provide a strong boost ($+2.3\%$ over real data), driven by parametric learning of disease-specific features. The combination of all three methods yields the best performance, validating our multi-pronged approach where each method addresses different scarcity scenarios.

\begin{table}[t]
  \centering
  \begin{minipage}[t]{0.49\textwidth}
    \centering
    \caption{\textbf{Classification accuracy per data type.}}\label{tab:ablation}
    \fontsize{7}{8.5}\selectfont
    \setlength\tabcolsep{3pt}
    \begin{tabular}{lcccc}
      \toprule
      Training Data & Mean & Light & Med. & Dark \\
      \midrule
      Real DDI only & 82.4 & 83.3 & 74.6 & 89.7 \\
      Healthy + Map & 81.2 & 82.1 & 78.4 & 82.9 \\
      Semantic only & 84.7 & 86.3 & 82.5 & 85.2 \\
      All cgDDI (Exp.~1) & \textbf{86.4} & \textbf{88.9} & \textbf{84.1} & 86.0 \\
      \bottomrule
    \end{tabular}
  \end{minipage}\hfill
  \begin{minipage}[t]{0.49\textwidth}
    \centering
    \caption{\textbf{Generative quality per skin tone.}}\label{tab:gen_fairness}
    \fontsize{7}{8.5}\selectfont
    \setlength\tabcolsep{3pt}
    \begin{tabular}{lccc}
      \toprule
      Skin Tone & FID $\downarrow$ & KID $\downarrow$ & LPIPS $\downarrow$ \\
      \midrule
      Light  & 103.45 & 0.039 & 0.715 \\
      Medium & 88.41  & 0.032 & 0.741 \\
      Dark   & 108.07 & 0.016 & 0.734 \\
      \midrule
      Max/Min & 1.22  & 2.41  & 1.04 \\
      \bottomrule
    \end{tabular}
  \end{minipage}
\end{table}

\subsection{Generative Fairness}
\label{sec:gen_fair}

We evaluate whether generation quality is equitable across skin tones via FID, KID, and LPIPS between cgDDI and held-out DDI images (Table~\ref{tab:gen_fairness}). All metrics demonstrate reasonable stability: FID dispersion is low ($\sigma{=}8.39$, max/min ratio $1.22$), LPIPS is near-identical across tones ($\sigma{=}0.011$). Each metric slightly favors a different tone (FID: Medium, KID: Dark, LPIPS: Light), indicating no systematic advantage for any population.

\section{Conclusion}
\label{sec:conclusion}

We present cgDDI, a hybrid generation framework that synthesizes fair and diverse dermatological imagery under extreme data constraints. By combining non-parametric lesion mapping with parametric generation and prior-preserving healthy priors, our method achieves state-of-the-art DDI classification (Accuracy $90.9\%$, up from $87.4\%$ prior best~\cite{aayushman-2024}) and leading fairness metrics (EOM: $86.6\%$, up from $69.6\%$). Cross-dataset validation on F17k confirms generalizability, and compatibility with SAMv3 enables scalability without expert masks. We release 266k+ synthetic images, code, and models to support equitable dermatological AI research. Limitations include the minimum $\sim$10 samples needed for high-quality parametric generation (which our non-parametric processing addresses to a degree) and the benefit of incorporating board-certified dermatologist review for quality assessment. Future directions include few-shot adaptation techniques and cross-disease transfer learning to further reduce data requirements.

\begin{credits}
\subsubsection{\discintname}
\new{The authors have no competing interests to declare that are relevant to the content of this article.}
\end{credits}

\bibliographystyle{splncs04}
\bibliography{main}

\end{document}